\documentclass[runningheads]{llncs}
\usepackage{graphicx}
\usepackage{cite}
\usepackage{amsmath}
\usepackage{bm, upgreek}
\usepackage{amssymb}
\usepackage{tabularx}
\usepackage{multirow}
\usepackage{booktabs}
\usepackage{adjustbox}

\usepackage{changepage} 

\usepackage[labelfont=bf]{caption} 
\usepackage{ragged2e}

\usepackage{array}
\newcolumntype{C}[1]{>{\centering\let\newline\\\arraybackslash\hspace{0pt}}m{#1}}

\usepackage[colorlinks]{hyperref}
\usepackage[svgnames]{xcolor}
\hypersetup{
colorlinks = True,
urlcolor = NavyBlue,
citecolor= NavyBlue,
linkcolor=black, 
breaklinks=true,
}
\usepackage{breakurl}  

\usepackage{makecell} 

\begin{document}
\title{Pre-trained Language Model for \\Biomedical Question Answering}
%
\titlerunning{Pre-trained Language Model for Biomedical Question Answering}
%

\author{Wonjin Yoon \and 
Jinhyuk Lee \and  Donghyeon Kim \and  \\
Minbyul Jeong \and  
Jaewoo Kang\thanks{To whom correspondence should be addressed.}}
\authorrunning{W. Yoon et al.}
%
\institute{Korea University, Seoul, Korea\\
    \email{\{wjyoon, jinhyuk\_lee, donghyeon, minbyuljeong, kangj\}@korea.ac.kr}\\
}
\maketitle              
\begin{abstract}
The recent success of question answering systems is largely attributed to pre-trained language models.
However, as language models are mostly pre-trained on general domain corpora such as Wikipedia, they often have difficulty in understanding biomedical questions.
In this paper, we investigate the performance of BioBERT, a pre-trained biomedical language model, in answering biomedical questions including factoid, list, and yes/no type questions. 
BioBERT uses almost the same structure across various question types and achieved the best performance in the 7th BioASQ Challenge (Task 7b, Phase B).
BioBERT pre-trained on SQuAD or SQuAD 2.0 easily outperformed previous state-of-the-art models. BioBERT obtains the best performance when it uses the appropriate pre-/post-processing strategies for questions, passages, and answers. 

\end{abstract}

\section{Introduction}
Language models pre-trained on large-scale text corpora achieve state-of-the-art performance in various natural language processing (NLP) tasks when fine-tuned on a given task~\cite{peters2018deep, radford2018improving, devlin2018bert}.
Language models have been shown to be highly effective in question answering (QA), and many current state-of-the-art QA models often rely on pre-trained language models \cite{talmor2019multiqa}.
However, as language models are mostly pre-trained on general domain corpora, they cannot be generalized to biomedical corpora \cite{zhu2018clinical, lee2019biobert, beltagy2019scibert, alsentzer2019publicly}.
Hence, similar to using Word2Vec for the biomedical domain \cite{pyysalo2013distributional}, a language model pre-trained on biomedical corpora is needed for building effective biomedical QA models.

Recently, Lee et al.~\cite{lee2019biobert} have proposed BioBERT which is a pre-trained language model trained on PubMed articles.
In three representative biomedical NLP (bioNLP) tasks including biomedical named entity recognition, relation extraction, and question answering, BioBERT outperforms most of the previous state-of-the-art models.
In previous works, models were used for a specific bioNLP task \cite{yoon2019collabonet, lim2018chemical, wiese2017neuralfull, rosso2018mindlab}. However, the structure of BioBERT allows a single model to be trained on different datasets and used for various tasks with slight modifications in the last layer.

In this paper, we investigate the effectiveness of BioBERT in biomedical question answering and report our results from the 7th BioASQ Challenge \cite{tsatsaronis2015overview, krithara2016results, nentidis2017results, nentidisetal2018results}.
Biomedical question answering has its own unique challenges.
First, the size of datasets is often very small (e.g., few thousands of samples in BioASQ) as the creation of biomedical question answering datasets is very expensive.
Second, there are various types of questions including factoid, list, and yes/no questions, which increase the complexity of the problem.

We leverage BioBERT to address these issues.
To mitigate the small size of datasets, we first fine-tune BioBERT on other large-scale extractive question answering datasets, and then fine-tune it on BioASQ datasets.
More specifically, we train BioBERT on SQuAD~\cite{rajpurkar2016squad} and SQuAD 2.0~\cite{rajpurkar2018know} for transfer learning.
Also, we modify the last layer of BioBERT so that it can be trained/tested on three different types of BioASQ questions.
This significantly reduces the cost of using biomedical question answering systems as the structure of BioBERT does not need to be modified based on the type of question.

The contributions of our paper are three fold: 1) We show that BioBERT pre-trained on general domain question answering corpora such as SQuAD largely improves the performance of biomedical question answering models. 
Wiese et al.~\cite{wiese2017neural} showed that pre-training on SQuAD helps improve performance. We test the performance of BioBERT pre-trained on both SQuAD and SQuAD 2.0.
2) With only simple modifications, BioBERT can be used for various biomedical question types including factoid, list, and yes/no questions.
BioBERT achieves the overall best performance on all five test batches of BioASQ 7b Phase B\footnote{\url{http://participants-area.bioasq.org/results/7b/phaseB/}}, and achieves state-of-the-art performance in BioASQ 6b Phase B.
3) We further analyze the role of pre- and post-processing in our system and show that different strategies often lead to different results.

The rest of our paper is organized as follows. First, we introduce our system based on BioBERT.
We describe task-specific layers of our system and various pre- and post-processing strategies.
We present the results of BioBERT on BioASQ 7b (Phase B), which were obtained using two different transfer learning strategies, and we further test BioBERT on BioASQ 6b on which our system was trained.

\section{Methods}
\begin{figure}[t]
\includegraphics[width=\textwidth]{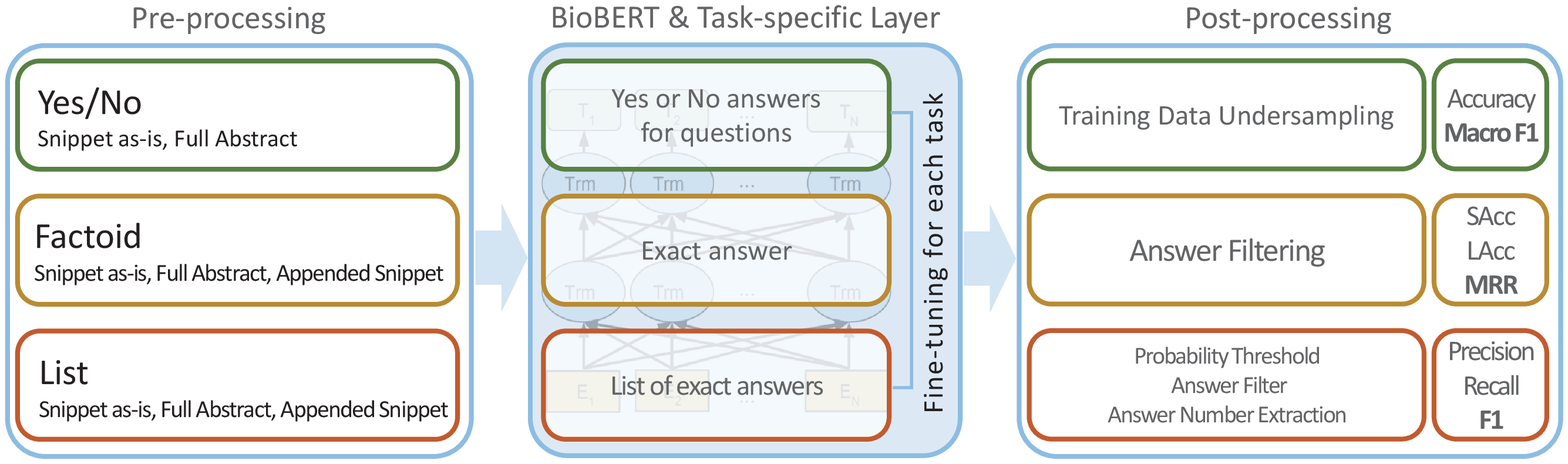}
\caption{Overview of our system.} \label{fig:overview} 
\end{figure}

In this section, we will briefly discuss BioBERT \footnote{The source code for BioBERT is available at \url{https://github.com/dmis-lab/biobert}.}\cite{lee2019biobert} and our modifications \footnote{The source code and pre-processed datasets are available at \url{https://github.com/dmis-lab/bioasq-biobert}.} for the BioASQ Challenge (Figure~\ref{fig:overview}). 

\subsection{BioBERT}
Word embeddings are crucial for various text mining systems since they represent semantic and syntactic features of words \cite{turian2010word, pyysalo2013distributional}. 
While traditional models use context-independent word embeddings, recently proposed models use contextualized word representations \cite{peters2018deep, radford2018improving, devlin2018bert}.
Among them, BERT~\cite{devlin2018bert}, which is built upon multi-layer bidirectional Transformers~\cite{vaswani2017attention}, achieved new state-of-the-art results on various NLP tasks including question answering.
BioBERT \cite{lee2019biobert} is the first domain-specific BERT based model pre-trained on PubMed abstracts and full texts. 
BioBERT outperforms BERT and other state-of-the-art models in bioNLP tasks such as biomedical named entity recognition, relation extraction, and question answering \cite{kim2019neural, sousa2019using}. 

An input representation of BioBERT for a given token is composed of the corresponding token, segment, and position embeddings. 
BioBERT utilizes WordPiece embeddings \cite{wu2016google} which use sub-word units to address the out-of-vocabulary (OOV) problem. 
Broken sub-word units are denoted by \#\# (e.g. \mbox{organoid} = \mbox{organ} + \mbox{\#\#iod}). 
Positional embeddings are learned during training and segment embeddings are used to mark the location of question and passage tokens in the input sequence. 
Following the design of BERT, a special token embedding for [CLS] was added to the beginning of every sequence to process yes/no type questions.

\begin{figure}[t]
\includegraphics[width=\textwidth]{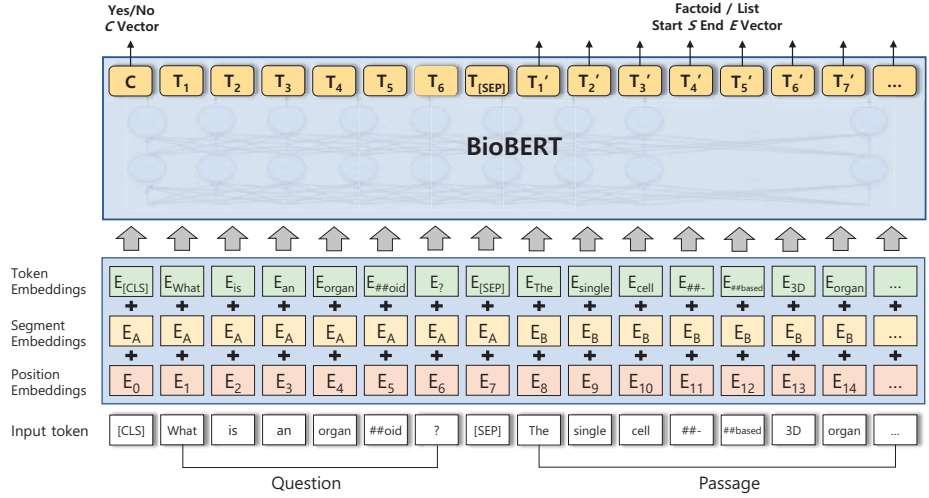}
\caption{Example of a single sequence (Question-Passage pair) processed by the BioBERT.} \label{fig:bert}
\end{figure}

\subsection{Task-specific layer}
The BioBERT model for QA is illustrated in Figure~\ref{fig:bert}.
Following the approach of BioBERT\cite{lee2019biobert}, a question and its corresponding passage are concatenated to form a single sequence which is marked by different segment embeddings.
The task-specific layer for factoid type questions and the layer for list type questions both utilize the output of the passage whereas the layer for yes/no type questions uses the output of the first [CLS] token.

\subsubsection{Factoid and List Questions}
In (Bio)BERT, the only additional trainable parameters needed for factoid and list type questions are the softmax layer for a linear transformation of hidden vectors from BioBERT. 
Following the notation used in the BERT study, we denote the trainable start vector as $S \in \mathbb{R}^H$ and the trainable end vector as $E \in \mathbb{R}^H$ where $H$ denotes the hidden size of BioBERT.
The probabilities of the $i$-th token being the start of the answer token and the $j$-th token being the end of the answer token can be calculated by the following equations: 
\[
P_i^{start} = \dfrac{e^{S \cdot T_i}}{\sum_{k} e^{S \cdot T_k}}, \text{ }
P_j^{end} = \dfrac{e^{E \cdot T_j}}{\sum_{k} e^{E \cdot T_k}} 
\]
where $T_l \in \mathbb{R}^H$ denotes $l$-th token representation from BioBERT and $\cdot$ denotes the dot product between two vectors. 

\subsubsection{Yes/no Questions}
We use the first [CLS] for the classification of yes/no questions.
Here, we denote the representation of the [CLS] token from BioBERT as $C \in \mathbb{R}^H$.
The parameter learned during training is a sigmoid layer consisting of $W \in \mathbb{R}^{H}$ which is used for binary classification.
The probability for the sequence to be ``yes'' is calculated using the following equation.
\[
P_{yes} = \frac{1}{1+e^{-CW}}
\]

\subsubsection{Loss}
For the factoid/list question layer, we minimize $Loss$ during training, which is defined below. $Loss$ is the arithmetic mean of the $Loss_{start}$ and $Loss_{end}$, which correspond to the negative log-likelihood for the correct start and end positions, respectively.
The ground truth start/end positions are denoted as $y_s$ for the start token, and $y_e$ for the end token. 
The losses are defined as follows:
\[
Loss_{start} = - \frac{1}{N}\sum_{k=1}^{N}{\log{P_{y_s}^{start, k}}}, \text{ }
Loss_{end} = - \frac{1}{N}\sum_{k=1}^{N}{\log{P_{y_e}^{end, k}}} 
\]
$$ Loss = (Loss_{Start} + Loss_{End})/2 $$
where $k$ iterates for a mini-batch of size $N$.\\
For yes/no questions, the binary cross entropy between probability $P_{yes}$ and the corresponding ground truth was used as the training loss.
\[
Loss = - (y_{yes} \log{P_{yes}} + (1-y_{yes}) \log{(1-P_{yes}))}
\]

\subsection{Pre-processing}
To solve the BioASQ 7b Phase B dataset as extractive question answering, the challenge datasets containing factoid and list type questions were converted into the format of the SQuAD datasets \cite{rajpurkar2016squad, rajpurkar2018know}. 
For yes/no type questions, we used 0/1 labels for each question-passage pair. 

The dataset in the SQuAD format consists of \textit{passages} and their respective question-answer sets. 
A passage is an article which contains answers or clues for answers and is denoted as the \textit{context} in the dataset.
The length of a passage varies from a sentence to a paragraph. 
An exact answer may or may not exist in the passage, depending on the task.
According to the rules of the BioASQ Challenge, all the factoid and list type questions should be answerable with the given passages \cite{tsatsaronis2015overview}.
An exact answer and its starting position are provided in the \textit{answers} field.
We used various sources including snippets and PubMed abstracts, as passages. 
Multiple passages attached to a single question were divided to form question-passage pairs, which increased the number of question-passage pairs. 
The predicted answers of the question-passage pairs which share the same question are later combined in the post-processing layer.  

Yes/no type questions are in the same format as the questions in the SQuAD dataset. However, binary answers are given to yes/no type questions, rather than answers selected based on their location in passages.
Instead of providing an exact answer and its starting position in the \textit{answers} field, we marked yes/no type questions using the strings ``yes'' or ``no''  and the Boolean values ``false'' and ``true'' in the \textit{is\_impossible} field.  
Since the distribution of yes/no answers in the training set is usually skewed, we undersampled the training data to balance the number of ``yes'' and ``no'' answers.

We used the following strategies for developing the datasets: \textit{Snippet as-is} Strategy, \textit{Full Abstract} Strategy, and \textit{Appended Snippet} Strategy.\vspace*{2pt} 
\begin{adjustwidth}{2em}{0pt}
\textit{\textbullet \hspace*{0.2em} Snippet as-is Strategy}
Using snippets in their original form is a basic method for filling passages.
The starting positions of exact answers indicate the positional offsets of exact matching words.
If a single snippet has more than one exact matching answer word, we form multiple question-passage pairs for the snippet.
\vspace*{2pt}\\
\textit{\textbullet \hspace*{0.2em} Full Abstract Strategy}
In the Full Abstract Strategy, we use an entire abstract, including the title of an article, as a passage. 
Full abstracts are retrieved from PubMed using their provided PMIDs.
The \textit{snippets} field of the original dataset is used to find the location of the correct answer.
First, we look for the given snippet (e.g., a sentence in a typical case) from the retrieved abstract.
Then, we search for the offset of the first exact matching words in the snippet, and add it to the offset of the snippet in the paragraph.
In this way, we can find a plausible location of the answer within the paragraph. 
\vspace*{2pt}\\
\textit{\textbullet \hspace*{0.2em} Appended Snippet Strategy}
The Appended Snippet Strategy is a compromise between using snippets as-is and full abstracts.
We first search a given snippet from an abstract and concatenate $N \in \mathbb{N}$ sentences before and after the given snippet, forming $2N + k$ sentences into a passage ($k$ denotes the number of sentences in a snippet, which is usually 1). 
\end{adjustwidth}

\subsection{Post-processing}
Since our pre-processing step involves dividing multiple passages with a same single question into multiple question-passage pairs, a single question can have multiple predicted answers.
The probabilities of predicted answers for question-passage pairs sharing the same question, were merged to form a single list of predicted answers and their probabilities for a question.
The answer candidate with the highest probability is considered as the final answer for a given factoid type question.
For list type questions, probability thresholding was the default method for providing answers.
Answer candidates with a probability higher than the threshold were included in the answer list.
However, a considerable number (28.6\% of BioASQ 6b list type questions) of list type questions contain the number of required answers. 
From the training example ``Please list 6 symptoms of Scarlet fever,"  we can extract the number 6 from the given question. 
We extracted the number provided in the question and used it to limit the length of the answer list for the question.
For questions that contain the number of answers, the extracted number of answers were yielded.  

For factoid and list type questions, we also filtered incomplete answers.
Answers with non-paired parenthesis were removed from the list of possible answers. 
Pairs of round brackets and commas at the beginning and end of answers were removed.

\section{Experimental Setup}
\subsection{Dataset}

For factoid and list type questions, exact answers are included in the given snippets, which is consistent with the extractive QA setting of the SQuAD \cite{rajpurkar2016squad} dataset. 
Only binary answers are provided for yes/no questions.
For each question, regardless of the question type, multiple snippets or documents are provided as corresponding passages.

The statistics of the BioASQ datasets are listed in Table~\ref{tab:dataset-stat}.
A list type question can have one or more than one answer; question-context pairs are made for every answer of a list type question.
In our pre-processing step, 3,722 question-context pairs were made from 779 factoid questions in the BioASQ 7b training set.
For yes/no questions, we undersampled the training data to balance the number of ``yes" and ``no" answers.

About 28.2\% of factoid type questions and 5.6\% of list type questions in the BioASQ 7b training set do not have an answer in their corresponding snippets.
We excluded unanswerable questions, following the approach of Wiese et al.\cite{wiese2017neuralfull}.

\begin{table}
\centering{}
\caption{Statistics of the BioASQ training set.}\label{tab:dataset-stat}
\begin{tabular}{@{}l|c|c|c@{}}
\toprule
\multirow{2}{*}{Question Type} & BioASQ & \multirow{2}{*}{\begin{tabular}[c]{@{}c@{}}\# of Questions in\\ original datasets\end{tabular}} & \multirow{2}{*}{\begin{tabular}[c]{@{}c@{}}\# of Pre-processed\\ question-passage pairs\end{tabular}} \\
 & Version &  &  \\ \midrule
Factoid & 6b & 618 & 3,121 \\
 & 7b & 779 & 3,722 \\ \midrule
List & 6b & 485 & 6,896 \\
 & 7b & 556 & 7,716 \\ \midrule
Yes/No & 6b & 612 & 5,921 \\
 & 7b & 745 & 6,676 \\ \bottomrule
\end{tabular}
\end{table}

\subsection{Training}
Our system is composed of BioBERT, task-specific layers, and a post-processing layer. 
The parameters of BioBERT and a task-specific layer are trainable.
Our training procedure starts with pre-training the system on the SQuAD dataset.
The trainable parameters for factoid and list type questions were pre-trained on the SQuAD 1.1 dataset, and the parameters for yes/no type questions were pre-trained on the SQuAD 2.0 dataset.
The pre-trained system is then fine-tuned on each task.

We tuned the hyperparameters on the BioASQ 4/5/6b training and test sets.
We used a probability threshold of 0.42 as one of the hyperparameters for list type questions. The probability threshold was decided using the tuning procedure. 


\section{Results \& Discussion}
In this section, we first report our results for the BioASQ 7b (Phase B) Challenge, which are shown in Table~\ref{tab:batch-results}. 
Please note that the results and ranks were obtained from the leaderboard of BioASQ 7b\cite{7bleaderboard}.
Then we evaluate our system and other competing systems on the validation set (BioASQ 6b).
The results are presented in Table~\ref{tab:6b-results}.
Finally, we investigate the performance gain due to the sub-structures of the system (Table~\ref{tab:ablation-pre-results} and Table~\ref{tab:ablation-post-results}).
Mean reciprocal rank (MRR) and mean average F-measure ($F_1$) were used as official evaluation metrics to measure the performance on factoid and list type questions from BioASQ, respectively.
We reported strict accuracy (SAcc), lenient accuracy (LAcc) and MRR for factoid questions and mean average precision, mean average recall, and mean average F1 score for list questions \footnote{For more details, please visit \url{http://participants-area.bioasq.org/Tasks/b/eval\_meas\_2018/}.}.
Since the label distribution was skewed, macro average F1 score was used as an evaluation metric for yes/no questions.

\subsection{Results on BioASQ 7b}
Our results on Task 7b (Phase B) of the BioASQ Challenge are reported in Table~\ref{tab:batch-results}.
Each participant can submit up to 5 systems per batch. We submitted 1 to 5 systems which use different combinations of pre- and post-processing strategies. 
We report the rankings and scores of our best performing system and those of other competing systems for each task in Table~\ref{tab:batch-results}. 
Competing systems are the best and second best systems, other than our system, from distinct participants.
Manually corrected gold-standard answers are not yet available at the time of writing; therefore, we report the scores based on the online leaderboard \footnote{The official results of the competition will be provided at \url{http://bioasq.org}.}.

\begin{table}
\caption{Batch results of the BioASQ 7b Challenge. We report the rank of the systems in parentheses. }\label{tab:batch-results} \vspace{5pt}
\begin{adjustbox}{max width=\textwidth}
\begin{tabular}{@{}c|l|c|l|c|l|c|c@{}}
\toprule
\multirow{2}{*}{Batch} & \multicolumn{2}{c|}{Yes/no} & \multicolumn{2}{c|}{Factoid} & \multicolumn{2}{c|}{List} & \multirow{2}{*}{\begin{tabular}[c]{@{}c@{}}\# of \\ Systems\end{tabular}} \\ \cmidrule(lr){2-7}
 & Participating system & Mac F1 & Participating system & MRR & Participating system & F1 &  \\ \toprule
\multirow{3}{*}{1} & (1) Ours & \textbf{67.12} & (1) Ours & \textbf{46.37} & (3) Ours & 30.51 & \multirow{3}{*}{17} \\
 & (2) auth-qa-1 & 53.97 & (2) BJUTNLPGroup & 34.83 & (1) Lab Zhu,Fudan Univer & \textbf{32.76} &  \\
 & (3) BioASQ\_Baseline & 47.27 & (3) auth-qa-1 & 27.78 & (4) auth-qa-1 & 25.94 &  \\ \midrule
\multirow{3}{*}{2} & (1) Ours & \textbf{83.31} & (1) Ours &\textbf{ 56.67} & (1) Ours & \textbf{47.32} & \multirow{3}{*}{21} \\
 & (2) auth-qa-1 & 62.96 & (3) QA1 & 40.33 & (3) LabZhu,FDU & 25.79 &  \\
 & (4) BioASQ\_Baseline & 42.58 & (4) transfer-learning & 32.67 & (5) auth-qa-1 & 23.21 &  \\ \midrule
\multirow{3}{*}{3} & (5) Ours & 46.23 & (6) Ours & 47.24 & (1) Ours & \textbf{32.98} & \multirow{3}{*}{24} \\
 & (1) unipi-quokka-QA-2 & \textbf{74.73} & (1) QA1/UNCC\_QA\_1 & \textbf{51.15} & (2) auth-qa-1 & 25.13 &  \\
 & (3) auth-qa-2 & 51.65 & (3) google-gold-input & 50.23 & (4) BioASQ\_Baseline & 22.75 &  \\ \midrule
\multirow{3}{*}{4} & (2) Ours & 79.28 & (1) Ours & \textbf{69.12} & (1) Ours & \textbf{46.04} & \multirow{3}{*}{36} \\
 & (1) unipi-quokka-QA-1 & \textbf{82.08} & (4) FACTOIDS/UNCC... & 61.03 & (2) google-gold-input-nq & 43.64 &  \\
 & (8) bioasq\_experiments & 58.01 & (9) google-gold-input & 54.95 & (9) LabZhu,FDU & 32.14 &  \\ \midrule
\multirow{3}{*}{5} & (1) Ours & \textbf{82.50} & (1) Ours & \textbf{36.38} & (1) Ours & \textbf{46.19} & \multirow{3}{*}{40} \\
 & (2) unipi-quokka-QA-5 & 79.39 & (3) BJUTNLPGroup & 33.81 & (6) google-gold-input-nq & 28.89 &  \\
 & (6) google-gold-input-ab & 69.41 & (4) UNCC\_QA\_1 & 33.05 & (7) UNCC\_* & 28.62 &  \\ \bottomrule
\end{tabular}
\end{adjustbox}
\end{table}

\subsection{Validating on the BioASQ 6b dataset}
We compared the performance of existing systems and our system on the BioASQ 6b dataset from the last year (2018), which is shown in Table~\ref{tab:6b-results}.
We micro averaged the scores from five experiments and reported the scores in Table~\ref{tab:6b-results}.
Similarly, the leaderboard scores of the best performing system for each batch were micro averaged and reported as the \textit{Best System} scores \cite{dimitriadis2019word, yang2016learning, peng2015fudan}.
Our system obtained much higher scores on the BioASQ 6b dataset than the top systems from leaderboard of BioASQ 6b Challenge.

\begin{table}
\centering{}
\caption{Performance comparison between existing systems and our system on the BioASQ 6b dataset (from last year). Note that our system obtained a 20\% to 60\% performance improvement over the best systems.  } \label{tab:6b-results} \vspace{5pt}
\begin{tabular}{@{}c| C{2.5cm} | C{2.5cm} | C{2.8cm} @{}}
\toprule
System & Factoid (MRR) & List (F1) & Yes/no (Macro F1) \\ \toprule
Best System & 27.84 \% & 27.21 \% & 62.05 \% \\ \midrule
Ours & \textbf{48.41} \% & \textbf{43.16} \% & \textbf{75.87} \% \\ \bottomrule
\end{tabular}
\end{table}

\subsubsection{Pre-training} In Table~\ref{tab:6b-modelcomparison}, we compare the performance of the pre-trained models.
BioBERT fine-tuned on the BioASQ 6b dataset outperformed BERT\textsubscript{BASE} fine-tuned on BioASQ in both factoid and list type questions.
BioBERT first pre-trained on SQuAD and then fine-tuned on BioASQ 6b obtained the best performance over other two experiments, demonstrating the effectiveness of pre-training BioBERT on SQuAD, a comprehensive and large-scale question answering corpus.

\begin{table}
\caption{Performance comparison between pre-trained models. } \label{tab:6b-modelcomparison}\vspace{5pt}
\begin{adjustbox}{max width=\textwidth}
\begin{tabular}{@{}l|c|c|c|c|c|c@{}}
\toprule
\multirow{2}{*}{Pre-trained models} & \multicolumn{3}{c|}{Factoid} & \multicolumn{3}{c}{List} \\ \cmidrule(l){2-7} 
 & SAcc & LAcc & MRR & Prec & Recall & F1 \\ \toprule
BERT\textsubscript{BASE}+BioASQ Finetune & 24.84\% & 36.03\% & 28.76\% & 42.41\% & 35.88\% & 35.37\% \\ \midrule
BioBERT+BioASQ Finetune & 34.16\% & 47.83\% & 39.64\% & 44.62\% & 39.49\% & 38.45\% \\ \midrule
BioBERT+SQuAD+BioASQ Finetune & \textbf{42.86\%} & \textbf{57.14\%} & \textbf{48.41\%} & \textbf{51.58\%} & \textbf{43.24\%} & \textbf{43.16\%} \\ \bottomrule
\end{tabular}\vspace{-10pt}
\end{adjustbox}
\end{table}

\subsubsection{Pre-/Post-processing}
The performance of our system is largely affected by how the data is pre-processed (Table~\ref{tab:ablation-pre-results}). 
However, the effectiveness of the pre-processing strategy varies depending on the type of question. 
For example, the Appended Snippet strategy and Full Abstract strategy obtained good performance on factoid questions, while the Snippet As-is strategy achieved the highest performance on list and yes/no type questions. 
Table~\ref{tab:ablation-post-results} shows the effect of post-processing on the performance of a system evaluated on list type questions.
In our study, both extracting the number of answers from questions and filtering predicted answers were effective.

\begin{table}
\centering{}
\caption{Performance comparison between pre-processing methods. Scores on the BioASQ 6b dataset.}\label{tab:ablation-pre-results}\vspace{5pt}
\begin{tabular}{@{}l|c|c|c|c|c|c|c@{}}
\toprule
\multirow{2}{*}{Strategy} & \multicolumn{3}{c|}{Factoid} & \multicolumn{3}{c|}{List} & Yes/no \\ \cmidrule(l){2-8} 
 & SAcc & LAcc & MRR  & Prec & Recall &  F1  & MacroF1\\
\toprule
Snippet  & 40.99 & 55.90 & 47.38 & \textbf{51.58} & \textbf{43.24} & \textbf{43.16} & 75.10\\
\midrule
 Full Abstract  & \textbf{42.86} & 57.14 & \textbf{48.41} & 42.66 & 32.58 & 33.52 & 66.76 \\
\midrule
 Appended Snippet & 39.75 & \textbf{58.39} & 48.00 & 44.04 & 41.26 & 39.36 & - \\
\bottomrule
\end{tabular}
\end{table}

\begin{table}
\centering{}
\caption{Ablation study on the post-processing methods. Scores for list type questions in the BioASQ 6b dataset.}\label{tab:ablation-post-results} \vspace{5pt}
\begin{tabular}{@{}l|C{1.3cm}|C{1.2cm}|C{1.2cm}@{}}
\toprule
 Strategy & Precision & Recall &  F1  \\
\toprule
 Baseline (Snippet) & \textbf{51.58} & 43.24 & \textbf{43.16} \\
\midrule
 Baseline without filter & 50.79 & 43.24 & 42.64 \\
\midrule
 Baseline without answer \# extraction & 50.01 & \textbf{44.32} & 42.58 \\
\bottomrule
\end{tabular}
\end{table}

\subsubsection{Ensemble}
Starting from test batch 4 of BioASQ 7b, we submitted model ensemble results as one of our systems. 
The performance gain of the model ensemble on our evaluation set was relatively small; the performance ranged from 0.2\% to 2\% depending on the task. 
The model ensemble improved the performance on factoid questions the most (2\% gain), but applying the model ensemble to list questions did not obtain higher performance than the  single model.
Although the model ensemble obtained high scores in the BioASQ 7b Challenge, it could only obtain the highest score on factoid type questions in batch 5.

\subsubsection{Qualitative Analysis}
In Table~\ref{tab:qual}, we show three predictions generated by our system on the BioASQ 6b factoid dataset.
Due to the space limitation, we show only small parts of a passage, which contain the answers (predicted answers might be contained in other parts of the passage).
We show the top five predictions generated by our system which can also be used for list type questions.
In the first example, our system successfully finds the answer and other plausible answers.
The second example shows that most of the predicted answers are correct and have only minor differences.
In the last example, we observe that the ground truth answer does not exist in the passage. Also, the predicted answers are indeed correct despite the incorrect annotation.

\begin{table}[!ht]
\caption{Predictions by our BioBERT based QA system on the BioASQ 6b factoid dataset}
\label{tab:qual}
\resizebox{\textwidth}{!}{%
\begin{tabular}{@{}cll@{}}
\toprule
No. & Type & Description \\ \midrule
1 & Question & What causes ``puffy hand syndrome?'' \\ \cmidrule(l){2-3} 
 & Passage & Puffy hand syndrome is a complication of intravenous drug abuse, \\
 & & which has no current available treatment.\\ \cmidrule(l){2-3} 
 & Ground Truth & ``intravenous drug abuse" \\ \cmidrule(l){2-3} 
 & Predicted Answer & ``intravenous drug abuse", \\
 & & ``drug addiction", \\
 & & ``Intravenous drug addiction", \\
 & & ``staphylococcal skin infection", \\
 & & ``major depression" \\ \toprule
2 & Question & In which syndrome is the RPS19 gene most frequently mutated? \\ \cmidrule(l){2-3} 
 & Passage & A transgenic mouse model demonstrates a dominant negative effect \\
 & & of a point mutation in the RPS19 gene associated \\
 & & with Diamond-Blackfan anemia. \\ \cmidrule(l){2-3} 
 & Ground Truth & ``Diamond-Blackfan Anemia", \\
 & & ``DBA" \\ \cmidrule(l){2-3} 
 & Predicted Answer & ``Diamond-Blackfan anemia", \\
 & & ``Diamond-Blackfan anemia (DBA)", \\
 & & ``DBA", \\
 & & ``Diamond Blackfan anemia", \\
 & & ``Diamond-Blackfan anemia. Diamond-Blackfan anemia" \\  \toprule
3 & Question & What protein is the most common cause of hereditary renal amyloidosis? \\ \cmidrule(l){2-3} 
 & Passage & We suspected amyloidosis with fibrinogen A alpha chain deposits, \\
 & & which is the most frequent cause of hereditary amyloidosis in Europe, \\
 & & with a glomerular preferential affectation. \\ \cmidrule(l){2-3} 
 & Ground Truth & ``Fibrinogen A Alpha protein" \\ \cmidrule(l){2-3} 
 & Predicted Answer & ``fibrinogen", \\
 & & ``fibrinogen alpha-chain. Variants of circulating fibrinogen", \\
 & & ``fibrinogen A alpha chain (FGA)", \\
 & & ``Fibrinogen A Alpha Chain Protein. Introduction: Fibrinogen", \\
 & & ``apolipoprotein AI" \\ \bottomrule
\end{tabular}%
}
\end{table}

The prediction result of list question from the BioASQ 6b is presented in Table~\ref{tab:qual-list}.
We found that our system is more likely to produce incorrect predictions on list questions than on factoid questions.
Our system internally outputs a list of predictions and the list is likely to include prediction with erroneous span. 
Even though incorrect prediction (``JBP'') with erroneous span has a lower probability than the true prediction (``JBP1'' and ``JBP2''), it can have considerable absolute probabilities.
On factoid questions, selecting a top one answer is required. 
Hence we can ignore incorrect prediction on factoid questions.
On the contrary, on list questions, prediction with erroneous span gets higher probability through merging predictions in post-processing step.
Since our model utilizes fixed threshold value, prediction with erroneous span is imperfect but achieved a higher possibility than the threshold.

\begin{table}[!ht]
\caption{Prediction by our BioBERT based QA system on the BioASQ 6b list dataset}
\label{tab:qual-list}
\resizebox{\textwidth}{!}{%
\begin{tabular}{@{}cll@{}}
\toprule
No. & Type & Description \\ \midrule
1 & Question & Which enzymes are responsible for base J creation in Trypanosoma brucei? \\ \cmidrule(l){2-3} 
 & Passage & JBP1 and JBP2 are two distinct thymidine hydroxylases involved in \\
 & & J biosynthesis in genomic DNA of African trypanosomes. \\ \cmidrule(l){3-3} 
 & & Here we discuss the regulation of hmU and base J formation in the  \\
 & & trypanosome genome by JGT and base J-binding protein.\\\cmidrule(l){2-3} 
 & Ground Truth & ``JBP1'', \\
 & & ``JBP2'', \\
 & & ``JGT''\\ \cmidrule(l){2-3} 
 & Predicted Answer & ``JBP1", \\
 & & ``JBP", \\
 & & ``thymidine hydroxylase", \\
 & & ``JGT", \\
 & & ``hmU", \\
 & & ``JBP2"\\ \bottomrule
\end{tabular}%
}
\end{table}

\section{Conclusion}
In this paper, we proposed BioBERT based QA system for the BioASQ biomedical question answering challenge.
As the size of the biomedical question answering dataset is very small, we leveraged pre-trained language models for biomedical domain which effectively exploit the knowledge from large biomedical corpora.
Also, while existing systems for the BioASQ challenge require different structures for different question types, our system uses almost the same structure for various question types.
By exploring various pre-/post-processing strategies, our BioBERT based system obtained the best performance in the 7th BioASQ Challenge, achieving state-of-the-art results on factoid, list, and yes/no type questions.
In future work, we plan to further systematically analyze the incorrect predictions of our systems, and develop biomedical QA systems that can eventually outperform humans.

\section*{Acknowledgements}
We appreciate Susan Kim for editing the manuscript.

\section*{Funding}
This work was funded by the National Research Foundation of Korea (NRF-2017R1A2A1A17069645, NRF-2016M3A9A7916996) and 
the National IT Industry Promotion Agency grant funded by the Ministry of Science and ICT and Ministry of Health and Welfare (NO. C1202-18-1001, Development Project of The Precision Medicine Hospital Information System (P-HIS)).

%
%
%
\bibliographystyle{splncs04}
\bibliography{bibgraphy}

\end{document}